\documentclass{article}

\usepackage{PRIMEarxiv}

\usepackage[utf8]{inputenc}
\usepackage[T1]{fontenc}
\usepackage{url}
\usepackage{booktabs}
\usepackage{tabularx}
\usepackage{algorithm}
\usepackage{algpseudocode}
\usepackage{amsfonts}
\usepackage{nicefrac}
\usepackage{microtype}
\usepackage{amsmath}
\usepackage{amsthm}
\usepackage{amssymb}
\usepackage{bm}
\usepackage{multirow}
\usepackage{caption}
\usepackage{subcaption}
\usepackage{graphicx}
\graphicspath{{Figures/}}
\usepackage{mathtools}
\usepackage{enumitem}
\usepackage{cite}
\usepackage[section]{placeins}

\usepackage{hyperref}
\hypersetup{
    colorlinks=true,
    citecolor=black,
    linkcolor=black,
    filecolor=black,
    urlcolor=black,
}

\newcommand{\R}{\mathbb{R}}

\numberwithin{equation}{section}

\title{Multi-Dimensional Training-Priority Weighting Based on Physical
Information Propagation Paths: A Unified Residual-Weighting Framework
for Physics-Informed Neural Networks}

\author{
  Zhangyi Lian \\
  College of Mechanical and Vehicle Engineering \\
  Chongqing University \\
  Chongqing 400030, China \\
  \And
  Xinda Dong \\
  State Key Laboratory of Tribology in Advanced Equipment \\
  Department of Mechanical Engineering \\
  Tsinghua University \\
  Beijing 100084, China \\
  \And
  Wenxuan Huo \\
  State Key Laboratory of Tribology in Advanced Equipment \\
  Department of Mechanical Engineering \\
  Tsinghua University \\
  Beijing 100084, China \\
  \And
  Weifeng Huang \\
  State Key Laboratory of Tribology in Advanced Equipment \\
  Department of Mechanical Engineering \\
  Tsinghua University \\
  Beijing 100084, China \\
  \And
  Greg Zhu \\
  State Key Laboratory of Tribology in Advanced Equipment \\
  Department of Mechanical Engineering \\
  Tsinghua University \\
  Beijing 100084, China \\
  \AND
  Qiang He$^*$ \\
  State Key Laboratory of Tribology in Advanced Equipment \\
  Department of Mechanical Engineering \\
  Tsinghua University \\
  Beijing 100084, China \\
  \texttt{heqiang@tsinghua.edu.cn}
}

\begin{document}
\maketitle

\begin{abstract}
Physics-informed neural networks (PINNs) have shown considerable promise for solving partial differential equations (PDEs); however, their synchronous optimization mechanism treats the residuals of different regions and different constraints with equal weight and in parallel, which is often inconsistent with the process by which the solution forms progressively ``from source to response'' along the physical information propagation path, thereby degrading training stability and solution accuracy. Existing causal training methods focus mainly on the temporal dimension and lack a unified characterization of the spatial and boundary dimensions. To address this, we define a unified class of training priorities according to the propagation path of physical information: along the PDE's physical information propagation path, premise regions should be learned before the regions that depend on them; the temporal, spatial, and boundary priorities are precisely instances of this principle on different propagation paths. On this basis, using neural tangent kernel (NTK) training dynamics, we theoretically analyze why the standard PINN does not obey this priority: its residual convergence order is governed by the NTK spectrum and is independent of the propagation path, and the spectral bias does not actively favor premise regions, so it does not establish a definite convergence order. Accordingly, we propose a unified multi-dimensional priority-constraint framework that, by partitioning the domain in order along the propagation path and constructing negative-exponential residual weights, converts the physical propagation order into a training priority at the loss level. For cases in which multiple priorities coexist, we introduce a directional compatibility coefficient to clarify the applicability boundary that ``orthogonal directions can be coupled multiplicatively in synergy, whereas coaxial opposite directions cannot.'' Multiple benchmark cases show that the proposed method consistently improves the convergence behavior and prediction accuracy of PINNs on problems with a clear propagation path or a constraint-dominated structure, without modifying the network architecture and with controllable additional computational cost.
\end{abstract}


\section{Introduction}
\label{sec:intro}

Physics-informed neural networks (PINNs), an emerging scientific-computing paradigm \cite{raissi2019physics,cuomo2022scientific,luo2025physics,karniadakis2021physics}, have shown broad application prospects in solving partial differential equations (PDEs) in recent years \cite{raissi2020hidden,cai2021physics,stiasny2023physics,zhang2022analyses,kissas2020machine}. The core advantage of PINNs lies in their ability to embed the PDE, the initial condition, and the boundary conditions uniformly into a single loss function, thereby avoiding the complex mesh generation and discretization-scheme construction required by traditional numerical methods. However, this unified loss form also implies an important assumption: that all physical constraints and all regions can be optimized synchronously and in parallel during training. For simple problems this assumption is usually acceptable; but for PDEs with pronounced temporal advancement, spatial transport, or boundary-dominated features, such synchronous optimization may not conform to the process by which the solution forms progressively along the physical information propagation path.

Existing studies have improved the training performance of PINNs mainly through network-architecture design \cite{wang2024nas,bu2021quadratic,zhao2024pinnsformer,jagtap2020adaptive}, loss-balancing strategies \cite{wang2022and,wang2021understanding,song2024loss,bischof2025multi,chen2018gradnorm}, domain decomposition and sequential learning \cite{kharazmi2021hp,moseley2023finite,shukla2021parallel,mattey2022novel,jagtap2020conservative,jagtap2020extended}, and adaptive weighting or sampling \cite{anagnostopoulos2024residual,zhang2025annealed,lu2021deepxde,daw2023mitigating,chen2025self,tang2023das}. These methods can alleviate problems such as gradient imbalance, local under-sampling, and insufficient fitting in complex regions, but most of them still retain a common premise: that the residuals of different regions, different time intervals, and different constraint sources can participate in optimization simultaneously. They therefore do not fundamentally answer a more basic question: should the training order of a PINN be consistent with the propagation path of physical information in the PDE?

From the perspective of physical-information propagation, Wang et al.~\cite{wang2024respecting} proposed a temporally causal PINN that, by imposing negative-exponential weights on time-blocked residuals, explicitly enforces the ``past $\to$ future'' training order and achieved good results on evolution-type problems. Subsequently, a series of related studies on causal modeling along the temporal dimension appeared \cite{li2024causality,guo2024tcas,qi2025rtc}. However, ordered propagation in PDEs does not exist only in the temporal dimension. Take the upwind scheme in computational fluid dynamics as an example: by preferentially using upstream information for discretization, it embodies the basic idea that ``the discretization process should be consistent with the direction of information propagation'' \cite{courant1952upwind}. In convection, transport, and wave problems, information propagates through space along characteristic directions; in boundary-dominated diffusion or steady-state problems, the boundary conditions act as constraint sources that transmit information from the boundary to the interior. It is thus evident that temporal causality is merely a special case of a more general ``training priority along the physical information propagation path.'' For an arbitrary PDE, physical information propagates from ``source'' to ``response'' along a propagation path $\mathcal{P}$ determined by the problem mechanism, and $\mathcal{P}$ induces on the regions an order in which ``premises precede dependents''---this is precisely the unified object that this paper aims to characterize, whereas existing work in the spatial dimension largely focuses on residual balancing \cite{mcclenny2023self,duan2025copinn,wu2023comprehensive} and seldom considers the direction of information propagation.

It should be noted that calling the above mechanism ``causal'' is not strictly accurate: for elliptic or strongly globally coupled problems, the solution has no unique propagation direction in space, and a strict causal relationship is difficult to define. To avoid conceptual overclaiming, this paper follows the line of thought of causal PINNs but strictly delimits the object of study as a class of ``training priorities based on the physical information propagation path,'' that is, assigning premise regions a higher learning priority than their dependent regions along $\mathcal{P}$. This formulation retains the physical motivation of ``following the physical propagation order'' while not relying on the existence of a strict causal relationship, and can therefore be naturally generalized to the temporal, spatial, and boundary dimensions.

Once this priority has been made explicit, a key question is whether the standard PINN obeys it. Using neural tangent kernel (NTK) training dynamics, this paper gives a negative answer: the residual convergence order of the standard PINN is governed by the NTK spectrum and is independent of the propagation path, and the spectral bias does not actively favor premise regions, so it does not establish a definite convergence order and therefore does not obey the priority based on the physical information propagation path. Accordingly, this paper proposes a unified multi-dimensional priority-weighting framework that interprets the temporal advancement, spatial propagation, and boundary-constraint transmission in a PDE uniformly as training priorities and, through ordered partitioning and negative-exponential residual weights, makes the PINN learn progressively in the manner of ``earlier regions are satisfied first, later regions are then released,'' which is subsequently verified on representative cases. The main contributions of this paper are summarized as follows:

\begin{itemize}
    \item We analyze the mechanism by which the standard PINN deviates from the physical propagation order: using NTK training dynamics, we show that the standard PINN does not actively obey this priority.
    \item On this basis, we propose a unified multi-dimensional priority-weighting framework that systematically extends the training priority from the temporal dimension to the spatial and boundary dimensions.
    \item We propose a directional-compatibility criterion for multi-dimensional priorities, clarifying the applicability boundary of ``orthogonal synergy, coaxial-opposite failure'' in coupling, and verify it on conflict cases.
\end{itemize}

The remainder of this paper is organized as follows. Section~\ref{sec:priority} defines a unified training priority according to the physical information propagation path and, using NTK training dynamics, analyzes why the standard PINN does not actively obey this priority; Section~\ref{sec:method} accordingly proposes a unified single- and multi-dimensional priority-weighting framework and provides support from the two perspectives of training dynamics and compatibility; Section~\ref{sec:experiments} verifies the method on a series of representative cases; and Section~\ref{sec:conclusion} concludes the paper and discusses future directions.

\section{Training Priority Based on Physical Information Propagation Path and Its Violation by the Standard PINN}
\label{sec:priority}

\subsection{Priorities Based on the Physical Information Propagation Path}
\label{sec:priority_def}

\textbf{The physical information propagation path.} In the physical system described by a partial differential equation, the solution in any region is not determined in isolation; rather, it is determined by information transmitted from other regions along a specific path by the governing equation and the definite-solution conditions. The upwind scheme in classical numerical analysis \cite{courant1952upwind,swanson1992central,leveque2002finite,toro2009riemann}, which preferentially uses upstream information for discretization, embodies the basic idea that ``the discretization process should be consistent with the direction of information propagation.'' Temporal evolution transmits information from earlier states to subsequent states; convection and waves transmit information from upstream to downstream along characteristic lines; and boundary conditions, as constraint sources, transmit information from the boundary to the interior. We refer to this directional information-transmission structure, determined by the physical mechanism, as the physical information propagation path, denoted $\mathcal{P}$. Along $\mathcal{P}$, the solution in each region depends on its ``premise'' regions that lie further ahead on the propagation path.

\textbf{Definition of the priority.} Along $\mathcal{P}$, the computational domain is partitioned in order into subregions $\{ D_{m}\}_{m = 1}^{M}$. If region $D_{i}$ transmits information along $\mathcal{P}$ to $D_{j}$---that is, the solution of $D_{j}$ depends on information from $D_{i}$ while $D_{i}$ does not depend on $D_{j}$---then $D_{i}$ is called a premise region of $D_{j}$, denoted
\begin{align}
    D_{i} \prec_{\mathcal{P}} D_{j}. \label{eq:prec}
\end{align}
The priority based on the physical information propagation path means the following: premise regions should be determined before their dependent regions, and the solution process advances progressively from ``source'' to ``response'' along $\mathcal{P}$, so premise regions should have a higher priority during learning. This priority is determined by the type of PDE and the definite-solution conditions and is independent of any specific solution algorithm. The physical origin of the priority in each of the three dimensions is explained below.

\begin{itemize}
    \item \textbf{Priority in the temporal dimension.} For an evolution-type PDE $u_{t} = \mathcal{N}[u]$, the system state evolves unidirectionally in time: the state at a subsequent instant is obtained by progressively advancing the state at an earlier instant through the equation, and information propagates along the time axis from the past to the future. Therefore, earlier time regions are the premises of later time regions, and the priority is ``past $\prec$ future.'' This is precisely the situation characterized by temporally causal training~\cite{wang2024respecting}---only after the solution at earlier instants has been sufficiently determined does the solution at later instants have a reliable basis.
    
    \item \textbf{Priority in the spatial dimension.} For convection-dominated or hyperbolic problems, when physical information is dominant along a certain direction in space along the characteristic direction, propagation can be approximately regarded as unidirectional. Take the upwind scheme in computational fluid dynamics~\cite{courant1952upwind} as an example: using only upstream information for discretization is precisely the embodiment of the rule that ``information propagates from upstream to downstream along the flow direction''---the state of a downstream region depends on the upstream, whereas the upstream should not be affected in reverse by the downstream. Therefore, upstream regions are the premises of downstream regions, and the priority is ``upstream $\prec$ downstream.''
    
    \item \textbf{Priority in the boundary dimension.} For diffusion-type or steady-state problems, there is no definite unidirectional propagation direction in the spatial interior, and the formation of the solution is mainly dominated by the boundary conditions. As constraint sources, the boundary conditions first constrain the solution shape in the boundary neighborhood and then transmit constraint information layer by layer toward the interior. Therefore, regions near the boundary are the premises of interior regions, and the priority is ``boundary $\prec$ interior.'' The initial condition can be regarded as a special case of this in the temporal dimension.
\end{itemize}

The three dimensions above are not mutually independent concepts but instances of the same principle on different propagation paths. For an arbitrary PDE, the physical information propagation path $\mathcal{P}$ induces a partial order $\prec_{\mathcal{P}}$ on the regions, and the priority always points from premise to dependent along $\mathcal{P}$. Accordingly, we uniformly define the three as a class of priorities based on the physical information propagation path. The three types of priority differ only in the specific form of $\mathcal{P}$ and share the same priority principle. This unified priority is the organizing concept of this paper: Section~\ref{sec:method} will explicitly inject it into the optimization process of the PINN through loss weighting.

\subsection{The Standard PINN and Its Synchronous-Optimization Assumption}
\label{sec:standard_pinn}

To motivate the subsequent analysis, this section reviews the basic form of the standard PINN and emphasizes the conceptual relationship between its synchronous-optimization assumption and the priority of Section~\ref{sec:priority_def}. Consider a space-time-dependent PDE problem
\begin{align}
    u_{t} + \mathcal{N}[u] = 0,\quad t \in [0,T],\ x \in \Omega;\quad u(0,x) = g(x);\quad \mathcal{B}[u] = 0 \label{eq:pde}
\end{align}
where $\mathcal{N}[\cdot]$ is a linear or nonlinear differential operator and $\mathcal{B}[\cdot]$ is a boundary operator. Following the approach of Raissi et al.~\cite{raissi2019physics}, the unknown solution $u(t,x)$ is approximated by a parameterized neural network $u_{\mathbf{\theta}}(t,x)$, which is trained by minimizing a composite loss:
\begin{align}
    \mathcal{L}(\mathbf{\theta}) = \lambda_{ic}\mathcal{L}_{ic}(\mathbf{\theta}) + \lambda_{bc}\mathcal{L}_{bc}(\mathbf{\theta}) + \lambda_{r}\mathcal{L}_{r}(\mathbf{\theta}) \label{eq:loss}
\end{align}
where the initial, boundary, and PDE residual losses are each measured by the mean-square residual at the sampling points; for example, the interior residual loss is
\begin{align}
    \mathcal{L}_{r}(\mathbf{\theta}) = \frac{1}{N_{r}}\sum_{i = 1}^{N_{r}}\left| \frac{\partial u_{\mathbf{\theta}}}{\partial t}(t_{r}^{i},x_{r}^{i}) + \mathcal{N}[u_{\mathbf{\theta}}](t_{r}^{i},x_{r}^{i}) \right|^{2}. \label{eq:loss_r}
\end{align}
The weights $\lambda_{ic},\lambda_{bc},\lambda_{r}$ are used to balance the various constraints, and the required derivatives are computed by automatic differentiation~\cite{griewank2008evaluating,baydin2018automatic}.

Conceptually, the composite loss~\eqref{eq:loss} sums the initial, boundary, and interior residuals synchronously with fixed weights and, in~\eqref{eq:loss_r}, aggregates the residuals at the individual sampling points with equal weight~\cite{wang2022and,jacot2018neural,lee2019wide}. This form implies a key assumption: that all physical constraints and all spatial and temporal regions can be optimized simultaneously and equivalently. However, this synchronous-optimization assumption contains no structure regarding the physical information propagation path---it treats premise regions and dependent regions alike, neither distinguishing their different roles in the formation of the solution nor possessing a mechanism of ``satisfy the premise first, then solve the dependent.'' In other words, by construction, the optimization objective of the standard PINN is agnostic to the unified priority defined in Section~\ref{sec:priority_def}: the formation of the solution should advance from source to response along $\mathcal{P}$, yet the loss regards the entire domain as a homogeneous target that can be advanced synchronously.

It should be emphasized that this conceptual ``agnosticism'' only indicates that the standard PINN does not actively obey the priority; it is not yet sufficient to assert that its training will actually deviate from the priority. The latter is a statement about training dynamics, which will be addressed in Section~\ref{sec:ntk} by means of a spectral analysis of the neural tangent kernel (NTK).

\subsection{Violation of the Unified Priority by the Standard PINN: An NTK Training-Dynamics Analysis}
\label{sec:ntk}

Starting from training dynamics, this section examines how the synchronous optimization of the standard PINN allocates learning resources among the regions during training and shows that this allocation bears no intrinsic relation to the unified priority of Section~\ref{sec:priority_def}. The analysis follows the NTK framework established by Wang et al.~\cite{wang2021understanding,wang2024respecting}: we first show that parameter sharing couples the residuals of the regions during training; then, by means of an NTK spectral decomposition, we obtain the core conclusion that ``the convergence rate is proportional to the NTK eigenvalue,'' thereby showing that regions with larger eigenvalues are learned earlier and more thoroughly by the network and thus have a higher effective priority; finally, we point out that the order in which the regions of the standard PINN are learned is governed entirely by the NTK spectrum, which is determined by the network architecture and the collocation geometry and is independent of the physical information propagation path, and therefore does not possess a mechanism to make training advance along the propagation path.

To focus on the priority issue, we first retain only the PDE residual term. Let
\begin{align}
    \mathcal{L}(\mathbf{\theta}) = \frac{1}{N}\sum_{i = 1}^{N}r_{i}^{2}(\mathbf{\theta}),\quad r_{i}(\mathbf{\theta}) = \frac{\partial u_{\mathbf{\theta}}}{\partial t}(\mathbf{z}_{i}) + \mathcal{N}[u_{\mathbf{\theta}}](\mathbf{z}_{i}) \label{eq:loss_simple}
\end{align}
where $\mathbf{z}_{i} = (t_{i},x_{i})$ is a sampling point and $r_{i}$ is its residual. Along the propagation path $\mathcal{P}$, consider a premise point and a dependent point: let $\mathbf{z}_{i} \in D_{i}$, $\mathbf{z}_{j} \in D_{j}$, and $D_{i} \prec_{\mathcal{P}} D_{j}$.

The gradient of the loss~\eqref{eq:loss_simple} with respect to the parameters is
\begin{align}
    \nabla_{\mathbf{\theta}}\mathcal{L}(\mathbf{\theta}) = \frac{2}{N}\sum_{i = 1}^{N}r_{i}(\mathbf{\theta})\,\nabla_{\mathbf{\theta}}r_{i}(\mathbf{\theta}). \label{eq:grad}
\end{align}

Since the PINN shares a single set of parameters $\mathbf{\theta}$ across all sampling points, the update direction is jointly determined by the residual-gradient terms of all sampling points. To characterize how the residuals at the individual points influence one another during training, we introduce the neural tangent kernel (NTK) of the residuals
\begin{align}
    \Theta(\mathbf{z}_{i},\mathbf{z}_{j}) = \left\langle \nabla_{\mathbf{\theta}}r_{i}(\mathbf{\theta}),\,\nabla_{\mathbf{\theta}}r_{j}(\mathbf{\theta}) \right\rangle. \label{eq:ntk}
\end{align}

Under the gradient flow $\dot{\mathbf{\theta}} = - \nabla_{\mathbf{\theta}}\mathcal{L}$, the evolution of the residual at the $i$-th point is
\begin{align}
    \frac{d r_{i}}{d t} = \nabla_{\mathbf{\theta}}r_{i} \cdot \dot{\mathbf{\theta}} = - \frac{2}{N}\sum_{j = 1}^{N}\Theta(\mathbf{z}_{i},\mathbf{z}_{j})\, r_{j}. \label{eq:res_evol}
\end{align}

Existing studies~\cite{wang2022and,jacot2018neural,lee2019wide} show that the off-diagonal terms $\Theta(\mathbf{z}_{i},\mathbf{z}_{j})$ ($i \neq j$) are generally nonzero. Hence $d r_{i}/d t$ depends on the residuals $r_{j}$ at all points, including those in the dependent region $D_{j}$ that lies further along the propagation path: the residual evolution of a premise region is not independent of its dependent region. This coupling causes the learning of the regions to constrain one another, but the coupling itself cannot yet determine which region is learned first; the following spectral analysis of Wang et al.~\cite{wang2021understanding} characterizes this.

Stacking all residuals into $\mathbf{r} = [r_{1},\ldots,r_{N}]^{\top}$,~\eqref{eq:res_evol} can be written in matrix form
\begin{align}
    \frac{d\mathbf{r}}{dt} = - \frac{2}{N}\, \mathbf{K}\, \mathbf{r},\quad \mathbf{K} = [\Theta(\mathbf{z}_{i},\mathbf{z}_{j})]_{i,j = 1}^{N} \label{eq:matrix_form}
\end{align}
where $\mathbf{K}$ is the neural tangent kernel matrix of the PINN and is positive semi-definite. Wang et al.~\cite{wang2021understanding} proved that as the network width tends to infinity, $\mathbf{K}$ converges in probability to a deterministic kernel $\mathbf{K}^{\ast}$ and remains approximately unchanged during training, i.e.,
\begin{align}
    \lim_{\text{width} \rightarrow \infty}\sup_{t \in [0,T]} \| \mathbf{K}(t) - \mathbf{K}(0) \|_{2} = 0,
\end{align}
giving rise to lazy training. In this limit,~\eqref{eq:matrix_form} is a linear ordinary differential equation. Performing a spectral decomposition of $\mathbf{K}^{\ast}$, $\mathbf{K}^{\ast} = \mathbf{Q}\mathbf{\Lambda}\mathbf{Q}^{\top}$ ($\mathbf{\Lambda} = \operatorname{diag}(\lambda_{1},\ldots,\lambda_{N})$, $\lambda_{i} \geq 0$), and letting $\widetilde{\mathbf{r}} = \mathbf{Q}^{\top}\mathbf{r}$ be the residual components in the eigenbasis,~\eqref{eq:matrix_form} decouples into
\begin{align}
    \widetilde{r}_{i}(t) = \widetilde{r}_{i}(0)\,\exp\left( - \frac{2}{N}\lambda_{i}t \right). \label{eq:decoupled}
\end{align}

Under the NTK lazy-training approximation, each residual mode of the standard PINN decays at a rate proportional to its NTK eigenvalue $\lambda_{i}$ (\eqref{eq:decoupled}): modes with large eigenvalues converge preferentially, whereas modes with small eigenvalues have their convergence markedly delayed. The convergence order of the standard PINN is therefore governed entirely by the NTK spectrum $\{\lambda_{i}\}$. In other words, the larger the NTK eigenvalue of a region, the more attention the network pays to that region during training and the earlier and more thoroughly it is learned, so the region has a higher effective priority; regions with smaller eigenvalues are markedly deferred.

It is worth noting that the NTK kernel that determines the convergence order is jointly determined by the network architecture and the collocation geometry, and its formation contains no structural information about the physical information propagation path: the spectrum itself neither distinguishes which regions are premises and which are dependents, nor possesses any mechanism that would naturally endow premise regions with larger eigenvalues. Consequently, there is no built-in alignment between the effective priority of the standard PINN determined by the spectrum and the unified priority of Section~\ref{sec:priority_def} defined by the physical information propagation path---whether the two coincide depends on the network and collocation of the specific problem, rather than being guaranteed by the algorithm itself. Wang et al.~\cite{wang2024respecting} further point out that the standard PINN exhibits an implicit bias: it tends to simultaneously drive down the residuals at all instants, and even those of all loss components, without distinguishing their order of precedence in the formation of the solution, which does not naturally coincide with the order of ``progressive formation from premise to dependent along the propagation path.''

It should be emphasized that this section does not assert that the eigenvalues of the later (dependent) regions must be larger than, or must differ from, those of the earlier (premise) regions---the specific size relationship of the eigenvalues among the regions depends on the problem, the network, and the collocation, and cannot be settled in the abstract here. What this section establishes is a mechanism-level conclusion: the standard PINN treats all regions alike in synchronous optimization, and the order in which they are learned is governed by the NTK spectrum, which is independent of the physical information propagation path; it therefore does not possess the directionality to make training advance along the propagation path and thus systematically fails to guarantee obeying the unified priority of Section~\ref{sec:priority_def}. As for whether, in a specific problem, the spectrum turns out to be roughly uniform across regions or happens to violate the prescribed priority, this will be illustrated in Section~\ref{sec:experiments} in combination with experimental results: by directly measuring the distribution of NTK eigenvalues during training, one can observe that the eigenvalues of the standard PINN are roughly uniform across the regions and do not form a clear decreasing structure along the ``premise--dependent'' direction, thereby corroborating the analysis of this section at the level of training dynamics.

This spectral perspective also indicates the direction of the remedy: Section~\ref{sec:method} will inject ordered priority weights $\mathbf{W}$ into the regions along the physical information propagation path and, by reshaping the NTK spectrum so that premise regions obtain larger effective eigenvalues, restore the directionality of training.

\section{Priority-Weighting Method Based on the Physical Information Propagation Path}
\label{sec:method}

\begin{figure}[H]
    \centering
    \captionsetup{name=Fig.,labelfont={normalfont},labelsep=period}
    \includegraphics[width=0.76\textwidth]{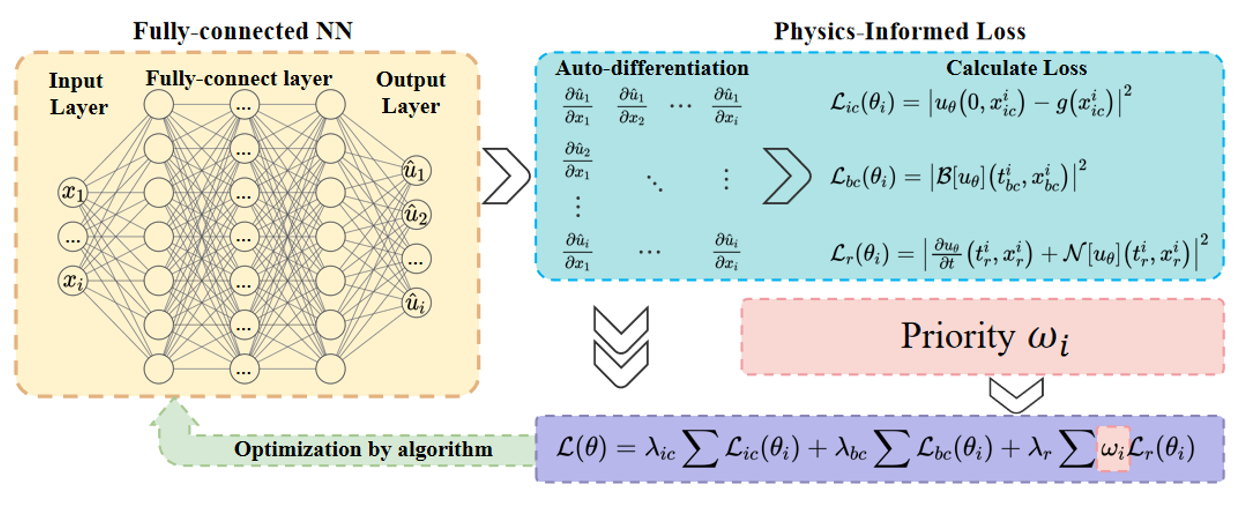}
    \caption{Structure of single-dimensional priority weighting. The priority weight $\omega_m$ of each subregion is constructed from the residuals of its premise regions via a negative exponential, so that the training advances progressively from the earlier regions to the later regions along the propagation path.}
    \label{fig:single_dim}
\end{figure}

This section explicitly realizes the priority defined in Section~\ref{sec:priority_def} as residual weighting at the loss level. Its core principle is that the residuals of the premise regions determine the weights of the subsequent dependent regions---only after the premise regions have been sufficiently fitted do the residuals of the dependent regions dominate the total loss, so that the optimization advances progressively from source to response along the physical information propagation path $\mathcal{P}$. This weighting does not change the network architecture or the way the residuals are computed by automatic differentiation; it acts only on the loss-aggregation layer (Fig.~\ref{fig:single_dim}).

\subsection{Single-Dimensional Priority-Weighting Algorithm}
\label{sec:single_dim}

\textbf{(1) Unified weighting form.} Along the propagation path $\mathcal{P}$, the computational domain is partitioned in order into subregions $\{ D_{m}\}_{m = 1}^{M}$ ($m$ increases from premise to dependent along $\mathcal{P}$). Before the PDE residuals enter the total loss, priority weights $\omega_{m}$ are introduced, and the composite loss is rewritten as
\begin{align}
    \mathcal{L}(\mathbf{\theta}) = \lambda_{ic}\sum_{i}\mathcal{L}_{ic}(\mathbf{\theta}) + \lambda_{bc}\sum_{i}\mathcal{L}_{bc}(\mathbf{\theta}) + \lambda_{r}\sum_{m = 1}^{M}\omega_{m}\,\mathcal{L}_{m}(\mathbf{\theta}) \label{eq:loss_unified}
\end{align}
where the residual loss of the $m$-th subregion is
\begin{align}
    \mathcal{L}_{m}(\mathbf{\theta}) = \frac{1}{N_{m}}\sum_{\mathbf{z}_{i} \in D_{m}}|r_{\mathbf{\theta}}(\mathbf{z}_{i})|^{2}. \label{eq:loss_subregion}
\end{align}
The priority weight is constructed from the cumulative residual of all premise regions of that region:
\begin{align}
    \omega_{m} = \exp\!\left( - \varepsilon\sum_{D_{k} \prec_{\mathcal{P}} D_{m}}\mathcal{L}_{k}(\mathbf{\theta}) \right) \label{eq:omega_def}
\end{align}
where $\varepsilon > 0$ controls the strength of the weight decay, and the summation runs over the premise regions that precede $D_{m}$ along $\mathcal{P}$. The single-dimensional priority-weighted residual is
\begin{align}
    \mathcal{L}^{c}(\mathbf{\theta}) = \frac{1}{M}\sum_{m = 1}^{M}\omega_{m}\,\mathcal{L}_{m}(\mathbf{\theta}). \label{eq:loss_weighted}
\end{align}
From~\eqref{eq:omega_def}, when the residual of a premise region is large, $\omega_{m}$ is suppressed and the subsequent regions are inhibited; as the premises are progressively satisfied and the cumulative residual becomes small, $\omega_{m} \to 1$ and the weights of the subsequent regions are released accordingly. The three dimensions differ only in the form of the premise relation $\prec_{\mathcal{P}}$, which is given separately below.

\textbf{(2) Single-direction weighting.} For the temporal dimension and part of the spatial dimension, $\mathcal{P}$ is a single direction, the blocks are ordered monotonically along that direction, and the premise regions are simply those with smaller indices $\{ D_{k}:k < m\}$.

Temporal causality (past $\prec$ future): the time domain is partitioned into ordered subintervals $[0,T] = \bigcup_{m = 1}^{M_{t}}I_{m}$,
\begin{align}
    \mathcal{L}_{t,m}(\mathbf{\theta}) = \frac{1}{N_{t,m}}\sum_{(t_{i},x_{i}) \in I_{m} \times \Omega}|r_{\mathbf{\theta}}(t_{i},x_{i})|^{2},\quad
    \omega_{t,m} = \exp\!\left( - \varepsilon_{t}\sum_{k = 1}^{m - 1}\mathcal{L}_{t,k}(\mathbf{\theta}) \right) \label{eq:temporal_weight}
\end{align}
\begin{align}
    \mathcal{L}^{c}_{t}(\mathbf{\theta}) = \frac{1}{M_{t}}\sum_{m = 1}^{M_{t}}\omega_{t,m}\,\mathcal{L}_{t,m}(\mathbf{\theta}). \label{eq:temporal_loss}
\end{align}

Spatial causality (upstream $\prec$ downstream): the spatial domain is partitioned along the dominant propagation direction into $\Omega = \bigcup_{m = 1}^{M_{s}}\Omega_{m}$,
\begin{align}
    \mathcal{L}_{s,m}(\mathbf{\theta}) = \frac{1}{N_{s,m}}\sum_{(t_{i},x_{i}) \in [0,T] \times \Omega_{m}}|r_{\mathbf{\theta}}(t_{i},x_{i})|^{2},\quad
    \omega_{s,m} = \exp\!\left( - \varepsilon_{s}\sum_{k = 1}^{m - 1}\mathcal{L}_{s,k}(\mathbf{\theta}) \right) \label{eq:spatial_weight}
\end{align}
\begin{align}
    \mathcal{L}^{c}_{s}(\mathbf{\theta}) = \frac{1}{M_{s}}\sum_{m = 1}^{M_{s}}\omega_{s,m}\,\mathcal{L}_{s,m}(\mathbf{\theta}). \label{eq:spatial_loss}
\end{align}
Both encode the single-direction transmission order of ``from source to response'' into the loss aggregation, making the upstream/early regions converge preferentially and the downstream/late regions be released thereafter.

\textbf{(3) Both-sides-toward-the-middle weighting.} When the problem has a boundary constraint on only one side, the boundary priority advances unidirectionally from the boundary toward the interior along that normal, and is handled in the same way as the single-direction weighting above. However, in realistic physical fields, boundary constraints usually exist simultaneously at both ends along the boundary normal $\Gamma_{b}$, both acting as constraint sources; and, as in the Burgers equation where the fluid also flows from both sides toward the middle, in such cases information is transmitted from the two boundaries toward the middle simultaneously---that is, the two boundaries are the premises and the middle region is the dependent.

If one were to order along a single linear sequence in this case, one side's boundary would be erroneously treated as the ``last dependent''; and if one forced both boundaries to serve simultaneously as premises and applied multiplicative coupling to the same normal, the two mutually opposite priorities would fall into the coaxial-opposite conflict ($\rho = -1$) described in Section~\ref{sec:multi_dim} and fail. For this reason, this paper adopts a partitioning strategy to realize ``both sides toward the middle'': along $\Gamma_{b}$, the computational domain is divided at the midline into two subregions $\Omega = \Omega_{1} \cup \Omega_{2}$, each dominated by its adjacent boundary, and single-direction weighting pointing from the boundary toward the midline is applied to each:
\begin{align}
    \mathcal{L}^{c}_{b}(\mathbf{\theta}) = \frac{1}{M_{1}}\sum_{m = 1}^{M_{1}}\omega_{1,m}\,\mathcal{L}_{1,m}(\mathbf{\theta}) + \frac{1}{M_{2}}\sum_{m = 1}^{M_{2}}\omega_{2,m}\,\mathcal{L}_{2,m}(\mathbf{\theta}) \label{eq:boundary_both}
\end{align}
\begin{align}
    \omega_{r,m} = \exp\!\left( - \varepsilon_{b}\sum_{k = 1}^{m - 1}\mathcal{L}_{r,k}(\mathbf{\theta}) \right),\quad r = 1,2 \label{eq:boundary_both_weight}
\end{align}
where the blocks of each subregion are ordered increasingly from its adjacent boundary toward the midline ($m = 1$ is closest to the boundary, $m = M_{r}$ is closest to the midline). Equivalently, the whole domain can be ordered uniformly by ``distance to the nearest boundary,'' taking the block closer to a boundary as the premise, yielding the same ``both-sides-toward-the-middle'' structure. This partitioning strategy localizes the two-sided boundary constraints, avoiding coaxial-direction conflict while simultaneously maintaining the dominant ``boundary $\rightarrow$ interior'' relation on both sides.

\FloatBarrier
\subsection{Directional Compatibility Coefficient and Multi-Dimensional Coupling}
\label{sec:multi_dim}

Practical PDE problems often contain multiple physical information propagation paths simultaneously. This section presents a coupling method for multi-dimensional priorities and uses a directional compatibility coefficient to characterize whether different priority directions can be compatibly coupled.

\textbf{(1) Multi-dimensional multiplicative coupling.} When multiple propagation paths are mutually independent, a separable multiplicative coupling is adopted. Let the set of priority directions involved in the coupling be $C$, and let the weight of the $q$-th direction be $\omega_{q}$; then the composite priority weight is
\begin{align}
    \omega_{\text{multi}} = \prod_{q \in C}\omega_{q}. \label{eq:omega_multi}
\end{align}
For a problem that simultaneously contains temporal advancement, spatial propagation, and boundary-constraint transmission,
\begin{align}
    \omega_{\text{multi}} = \omega_{t}\,\omega_{s}\,\omega_{b}. \label{eq:omega_multi_tsb}
\end{align}
The corresponding multi-dimensional priority-weighted residual is
\begin{align}
    \mathcal{L}_{\text{multi}}^{c}(\mathbf{\theta}) = \frac{1}{|M|}\sum_{m \in M}\omega_{m}\,\mathcal{L}_{m}(\mathbf{\theta}) \label{eq:loss_multi}
\end{align}
where $m$ is the multi-dimensional block index and $\omega_{m}$ is obtained from the product of the single-dimensional weights. This product form can be understood as ``multiple preconditions satisfied simultaneously'': a region obtains a high training priority only when the premise dependencies in all relevant directions are satisfied. The block-by-block computational procedure for the multi-dimensional blocks and weights is shown in Algorithm~\ref{alg:coupled_priority}.

\begin{algorithm}[htbp]
\small
\caption{Coupled priority algorithm.}
\label{alg:coupled_priority}
\begin{algorithmic}[1]
\Require points in single batchsize
\Ensure special weights and residuals
\State Partition the points into $K_y$ chunks based on $y$ in dependence order
\State Calculate the total residual $\lambda_{yi}$ for each block
\Comment{Process $y$-axis direction dependence}
\For{$i = 1$ to $K_y$}
    \State Compute and log the $Y$-axis weights by: $w_{yi} = \exp(-\epsilon\sum_{j=0}^{i-1}\lambda_{yj})$
    \Comment{Process $x$-axis direction dependence}
    \State Fetch points in $k_y$ chunk
    \State Partition the points into $K_x$ chunks based on $x$ in dependence order
    \State Calculate the total residual $\lambda_{xi}$ for each block
    \For{$j = 1$ to $K_x$}
        \State Compute and log the $X$-axis weights by: $w_{xj} = \exp(-\epsilon\sum_{k=0}^{j-1}\lambda_{xk})$
        \State Multiply $w_{yi}$ by $w_{xj}$ to obtain $w_{x_iy_i}$
    \EndFor
    \State Concatenate $w_{x_iy_i}$ to $w_i$
\EndFor
\State Concatenate $w_i$ to $w$
\end{algorithmic}
\end{algorithm}

\textbf{(2) Directional compatibility coefficient.} Whether multiplicative coupling is effective depends on whether the two priority directions involved in the coupling are compatible. Let the transmission directions of the two priorities be $\mathbf{d}_{1},\mathbf{d}_{2} \in \R^{d}$, and define the directional compatibility coefficient
\begin{align}
    \rho(\mathbf{d}_{1},\mathbf{d}_{2}) = \mathbf{d}_{1} \cdot \mathbf{d}_{2} \in [-1,1]. \label{eq:rho_def}
\end{align}
Accordingly, multi-dimensional coupling is divided into three types (Table~\ref{tab:compatibility}). It should be noted that here ``direction'' refers to the direction from the start to the end of the priority ordering, rather than the physical flux direction; since in most problems $\rho$ is $-1$, $0$, or $1$, the case in which $\rho$ takes other values is not discussed for now.

\begin{table}[htbp]
\centering
\small
\caption{Compatibility classification of priority directions.}
\label{tab:compatibility}
\begin{tabularx}{\textwidth}{@{}lXXl@{}}
\toprule
Compatibility type & $\rho$ & Meaning & Typical situation \\
\midrule
Orthogonal-compatible & $\rho=0$ & Two directions act on different coordinate axes & time $\times$ space, time $\times$ boundary \\
Co-directional-compatible & $\rho=1$ & Two directions partly coincide and reinforce & coaxial co-directional propagation \\
Conflicting & $\rho=-1$ & Two directions are opposite and suppress each other & coaxial-opposite spatial priorities \\
\bottomrule
\end{tabularx}
\end{table}

\textbf{(3) Compatibility analysis.}

Orthogonal case ($\rho = 0$): the two directions act on different index dimensions, and the weight on the composite block $(m,n)$ is
\begin{align}
    \omega_{m,n} = \omega_{m}^{(1)}\,\omega_{n}^{(2)} = \exp\!\left( - \varepsilon_{1}\sum_{k < m}\mathcal{L}_{k}^{(1)} - \varepsilon_{2}\sum_{l < n}\mathcal{L}_{l}^{(2)} \right). \label{eq:omega_orthogonal}
\end{align}
Since the two summations act on different index sets, the starting subblock $(1,1)$, on which ``the premises of both directions are satisfied,'' obtains a high weight $\omega_{m,n} \to 1$ early in training, the progressive structure is preserved, and the multiplicative coupling forms synergy.

Coaxial-opposite case ($\rho = -1$): for two opposing priorities along the same coordinate direction $x$, with A ordered along $+x$ as $\omega_{m}^{A} = \exp(-\varepsilon\sum_{k < m}\mathcal{L}_{k})$ and B ordered along $-x$ as $\omega_{m}^{B} = \exp(-\varepsilon\sum_{k > m}\mathcal{L}_{k})$, their product is
\begin{align}
    \omega_{m}^{AB} = \omega_{m}^{A}\,\omega_{m}^{B} = \exp\!\left( - \varepsilon\sum_{k \neq m}\mathcal{L}_{k} \right). \label{eq:omega_conflict}
\end{align}
The weight of any block $m$ is suppressed by ``the cumulative residual of all blocks except itself'': there is no starting block that can preferentially obtain a high weight, all $\omega_{m}^{AB} \to 0$ early in training, and the progressive structure is completely lost. This explains the phenomenon in the Poiseuille and Couette cases in which coaxial-conflict products cause the error to jump to the order of $10^{-1}$; it is also precisely the reason why the two-sided boundaries in Section~\ref{sec:single_dim} must adopt partitioning rather than a product---the two boundaries are essentially a pair of $\rho = -1$ priorities, and partitioning localizes them to avoid conflict.

\textbf{(4) Coupling strategy.} Combining the above analysis, multi-dimensional coupling is selected according to compatibility:
\begin{itemize}
    \item If $\rho = 0$: multiplicative coupling, forming synergy.
    \item If $\rho = -1$: select the dominant priority on that axis and discard the conflicting direction.
    \item If $\rho = 1$: keeping either direction suffices, and the multiplicative coupling is algebraically equivalent to doubling the tolerance.
\end{itemize}

\FloatBarrier
\subsection{Implementation and Stability Considerations}
\label{sec:dynamics}

This subsection discusses several practical aspects of the proposed priority-weighting method from the perspectives of computational complexity, training-dynamics interpretation, hyperparameter selection, stability under directional conflict, and generality.

\textbf{(1) Computational complexity.} The method acts only on the loss function; it does not change the network architecture nor introduce additional high-order derivatives. In the single-dimensional case, the weights consist only of the accumulation, scaling, and exponentiation of block-level residuals, and the additional overhead is negligible; in the multi-dimensional coupling case, the weights extend to multi-dimensional tensors, involving additional storage, broadcasting, and products, so the training cost increases somewhat but overall remains significantly lower than the main computational cost of the network's forward and backward propagation.

\textbf{(2) Spectral reshaping and asymptotic consistency.} From the NTK training-dynamics perspective, the priority-weighted framework of Section~\ref{sec:single_dim} can be interpreted as reshaping the effective NTK spectrum. Let the weight matrix be $\mathbf{W} = \operatorname{diag}(\omega_{1}\mathbf{I}_{N_{1}},\ldots,\omega_{M}\mathbf{I}_{N_{M}})$, where $\mathbf{I}_{N_{m}}$ is the $N_{m} \times N_{m}$ identity matrix of the $m$-th subregion. The residual-evolution equation of the priority-weighted method is
\begin{align}
    \frac{d\mathbf{r}}{dt} = - \frac{2}{N}\,\mathbf{K}\,\mathbf{W}\,\mathbf{r}. \label{eq:weighted_dynamics}
\end{align}
Compared to the standard PINN~\eqref{eq:matrix_form},~\eqref{eq:weighted_dynamics} replaces $\mathbf{K}$ with $\mathbf{K} \to \mathbf{K}\mathbf{W}$. The ordered weight matrix $\omega_{m}$ along the propagation path reduces the effective eigenvalues of the dependent regions relative to the premise regions, thereby reshaping the effective NTK spectrum. As a result, the premise regions have larger effective eigenvalues and are learned earlier and more thoroughly, while the dependent regions are deferred until the premise regions have been satisfied.

\textbf{(3) Matching $\varepsilon$ to the residual scale.} The priority weight depends exponentially on the cumulative residual, and its numerical range significantly affects the weight distribution: when the residual scale is large, the weight decays rapidly, so the subsequent regions participate insufficiently early in training; when the residual scale is small, the weight differences between regions are weakened and the weighting effect diminishes. Assuming the cumulative residual of the preceding regions decreases as $\mathcal{L}_{k}(t) \sim C t^{-\alpha}$, the release time of the $m$-th region (the instant at which $\omega_{m} = 1/e$) is approximately $T_{m} \sim (\varepsilon m C)^{1/\alpha}$. When $\varepsilon$ is too large, $T_{M}$ exceeds the total number of training steps and the subsequent regions are always inhibited; when $\varepsilon$ is too small, the method degenerates into the standard PINN. A reasonable $\varepsilon$ should make $T_{M} \lesssim T_{\text{total}}$, i.e., the last subregion completes its release before training ends, where $T_{\text{total}}$ is the total training-iteration budget. Additionally, the multiplicative coupling of mutually independent priorities ($\rho = 0$) further reduces the effective eigenvalues of the dependent regions in the multiplication direction via the weight matrix $\mathbf{W}$, thereby strengthening the propagation priority.

\textbf{(4) Stability under conflict.} When multiple priorities coexist and there is a conflict on a certain coordinate axis ($\rho < 0$), a more robust approach is to select, on that axis, the priority corresponding to the dominant physical mechanism. This strategy does not mean restoring a unique physical evolution path; rather, it provides the PINN with a more stable and more executable optimization trajectory, avoiding the disruption of the progressive structure caused by direct competition between opposite priorities in the same direction. The conflicting case ($\rho < 0$) causes the eigenvalues of the same region to be suppressed simultaneously by the opposite-direction priorities, so that the region cannot obtain a sufficient effective eigenvalue and is thus difficult to train.

\textbf{(5) Generality.} Since the strategy relies only on reweighting at the loss level and is decoupled from the network architecture, its idea can in principle be combined with fully connected networks, Fourier-feature networks, or Transformer-based PINN variants, and can be embedded into existing frameworks without modifying the network topology.

\section{Numerical Experiments and Analysis of Results}
\label{sec:experiments}

To evaluate the performance of the proposed multi-dimensional priority-weighting framework under different physical-information-transmission structures, this paper selects a set of representative PDE cases for numerical experiments. The selected problems cover typical situations such as temporal evolution, spatial propagation, boundary driving, multi-dimensional coupling, and directional conflict, and include Stokes, Heat Conduction, Burgers, Allen--Cahn, Poiseuille Flow, and Couette Flow, corresponding respectively to empirical evaluations under single-dimensional priority weighting, multi-dimensional priority coupling, and conflicting-priority structures, and used to verify whether the unified priority-weighting framework proposed in Chapter~3 can make the PINN learn, in a more orderly manner, the physical information carried by different propagation directions and constraint sources.

To ensure comparability among the different methods, unless otherwise stated, all cases adopt the same basic training configuration (Adam optimizer, grad-norm residual-balancing method; detailed parameters are given in the appendix). For comparison, this paper introduces a neural-tangent-kernel-based weighting method (replacing grad norm with the NTK algorithm)~\cite{wang2022and} and a residual-based adaptive-sampling method (RAR)~\cite{lu2021deepxde}, and adopts the relative $L_{2}$ error as the main accuracy metric.

\subsection{Experimental Evaluation of Single-Dimensional Priority Weighting}
\label{sec:exp_single}

\subsubsection{Spatial Priority Weighting: Stokes Flow Past a Cylinder}

To verify the effectiveness of spatial priority weighting in the unified priority framework, this paper selects the Stokes equations for steady flow of an incompressible viscous fluid at very low Reynolds numbers as a typical case. The governing equations are
\begin{align}
    -\nabla p + \mu\nabla^{2}\mathbf{u} + \mathbf{f} = 0,\quad \nabla\cdot\mathbf{u} = 0, \label{eq:stokes}
\end{align}
where $\mathbf{u} = (u,v)$ is the velocity field, $p$ is the pressure field, $\mu$ is the dynamic viscosity, and $\mathbf{f}$ is the body-force term; the two equations above express momentum conservation and the incompressibility constraint, respectively. In this experiment, a cylindrical obstacle is placed in a rectangular computational domain, a parabolic velocity profile is imposed on the inlet boundary, a fully developed flow or natural outflow condition is adopted on the outlet boundary, and no-slip boundary conditions are imposed on the solid walls and the cylinder surface.

In this flow-past-a-cylinder configuration, the inlet driving, the open outlet boundary, and the cylindrical obstacle together form a fairly definite mainstream direction, so that the error distribution may exhibit regional differences along the flow direction during training. Accordingly, this paper partitions the computational domain into several ordered spatial subregions along the mainstream direction and uses spatial priority weighting to construct the residual weights, so that the fitting state of the upstream regions modulates the contribution of the downstream regions' residuals to the total loss.

\begin{figure}[H]
    \centering
    \captionsetup{name=Fig.,labelfont={normalfont},labelsep=period}
    \includegraphics[width=0.76\textwidth]{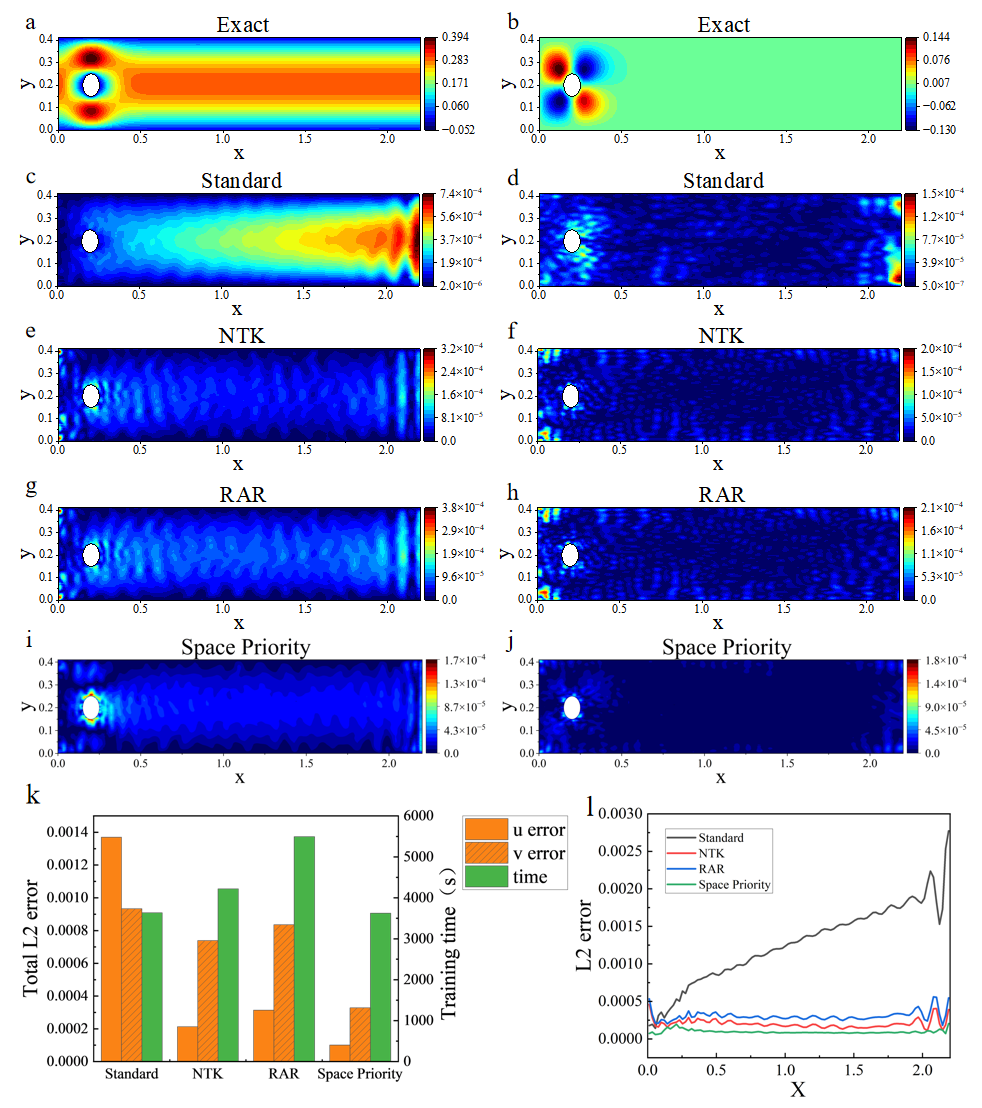}
    \caption{Reference solution of the Stokes flow-past-a-cylinder case, the predicted error distributions of different methods, the global $L_{2}$ error, the training time, and the error distribution along the $x$-axis.}
    \label{fig:stokes}
\end{figure}

As can be seen from Fig.~\ref{fig:stokes}(c, d), the error of the standard PINN is concentrated mainly behind the cylindrical obstacle and in the downstream region. Under the training scheme of global synchronous residual aggregation, although the model is simultaneously constrained by the PDE residuals throughout the computational domain, it still exhibits fairly obvious under-fitting of the local flow structure formed after the obstacle's disturbance, together with a phenomenon of error accumulation in space; the residuals of the downstream region interfere with the overall parameter update, thereby weakening the model's ability to progressively form a stable flow-field representation along the mainstream direction.

Compared with the standard PINN, the spatial-priority-weighted PINN shows a marked reduction in error behind the cylinder and in the downstream region, and the error-distribution contour is more uniform overall. This further indicates that priority weighting does not merely change the final global error value, but transforms the error-correction process from the global synchronous competition of the standard PINN into a structured optimization process that unfolds progressively along the mainstream direction, thereby effectively improving the error distribution in the downstream wake region.

On the basis of corroborating the above changes in the error distribution, the quantitative results in Fig.~\ref{fig:stokes}(g, h) further provide a lateral comparison of the spatial-priority-weighted PINN with the RAR and NTK weighting methods. Although all three methods attempt to improve the training of the standard PINN, their mechanisms differ markedly: RAR mainly strengthens local fitting by increasing the sampling density in high-residual regions; NTK weighting adjusts the influence of different residual terms on the parameter update from the perspective of training dynamics and loss-contribution balancing. By contrast, the training accuracy of spatial priority weighting is higher than that of the other two methods, and its improvement is reflected not only in the reduction of the relative $L_{2}$ error but also in the shift of the error distribution from being concentrated in the cylinder wake to a smoother and more balanced spatial form.

In summary, in this steady flow-past-a-cylinder problem, the spatial priority weighting constructed along the mainstream direction can adjust the training priority of the PINN for regions with different flow orientations, making the model coordinate the learning among the upstream region, the obstacle neighborhood, and the downstream wake region in a more orderly manner, thereby alleviating the local error concentration downstream and improving the overall prediction accuracy.

\FloatBarrier
\subsubsection{Boundary Priority Weighting: Heat Conduction}

To evaluate the role of boundary priority weighting in diffusion-type problems, this paper selects the one-dimensional unsteady heat-conduction equation as a representative case. Consider an infinite plate of thickness 2 undergoing transient heat conduction in the absence of an internal heat source. Since the temperature gradients along the length and width directions are negligible, the problem can be reduced to a one-dimensional heat-conduction model. Using the symmetry of the temperature field about the mid-plane of the plate, this paper considers only the half-domain $x \in [0,1]$, and its governing equation can be written as:
\begin{align}
    \frac{\partial t}{\partial \tau} = a\frac{\partial^{2} t}{\partial x^{2}}, \label{eq:heat}
\end{align}
where $t$ denotes the dimensionless temperature, $\tau$ denotes the dimensionless time, and $a$ is the thermal diffusivity. Its initial condition is:
\begin{align}
    t = 1,\quad \tau = 0, \label{eq:heat_ic}
\end{align}
The boundary conditions are:
\begin{align}
    \frac{\partial t}{\partial x} = 0,\quad x = 0, \label{eq:heat_bc1}
\end{align}
\begin{align}
    -\lambda\frac{\partial t}{\partial x} = h t,\quad x = 1. \label{eq:heat_bc2}
\end{align}

In this heat-conduction problem, heat diffusion does not correspond to a clear unidirectional spatial propagation path; instead, the heat-exchange boundary, as a continuously acting constraint source, plays an important anchoring role in the formation of the interior temperature field. Based on this physical feature, this paper partitions the computational domain in order along the boundary-normal direction and adopts a boundary-priority-weighting form, so that the regions near the heat-exchange boundary have a higher optimization priority early in training, while the training contribution of the interior regions is released progressively as the fitting state of the boundary neighborhood improves.

\begin{figure}[H]
    \centering
    \captionsetup{name=Fig.,labelfont={normalfont},labelsep=period}
    \includegraphics[width=0.76\textwidth]{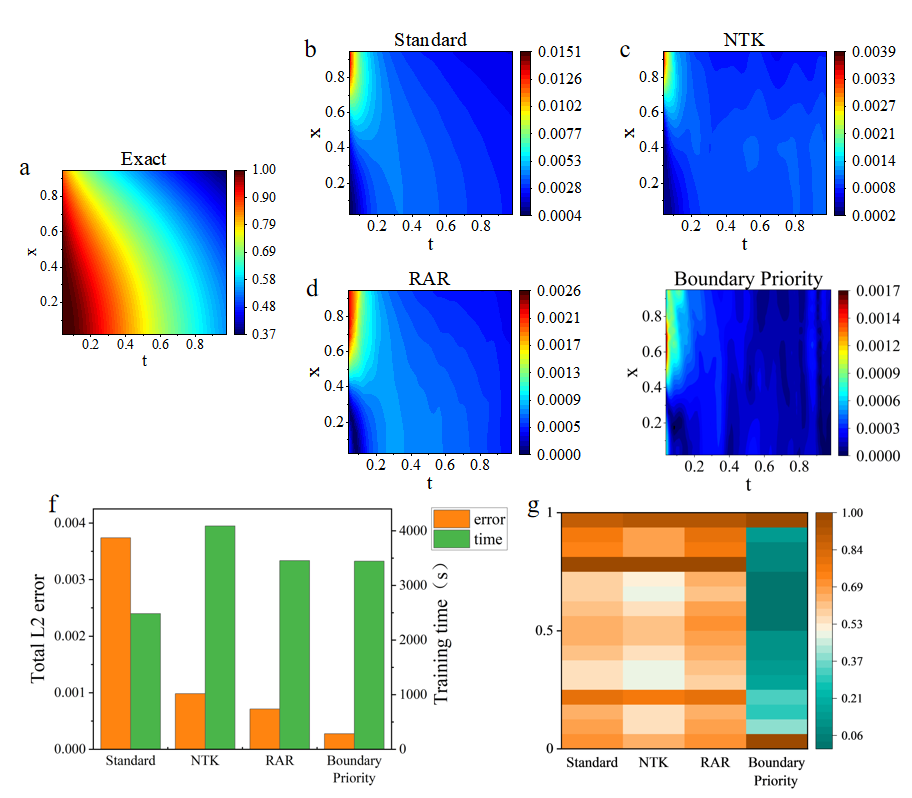}
    \caption{Reference solution of the Heat Conduction case, the error distributions of different methods, the global $L_{2}$ error, the training time, and the normalized distribution of the block-wise NTK eigenvalues along the $x$-axis in the early training stage (10,000 steps).}
    \label{fig:heat}
\end{figure}

From the error distribution in Fig.~\ref{fig:heat}(b), the error of the standard PINN is concentrated mainly near the heat-exchange boundary and extends toward the interior region along the boundary normal. Fig.~\ref{fig:heat}(c) shows that after boundary priority weighting is introduced, the error concentration near the boundary is markedly weakened and the error transition from the boundary normal toward the interior region is smoother.

Fig.~\ref{fig:heat}(f) compares the different algorithms; boundary priority weighting outperforms the standard PINN, RAR, and NTK weighting methods in terms of the overall relative $L_{2}$ error. Compared with NTK and RAR, boundary priority weighting constructs the residual weights directly around the heat-exchange boundary, reducing the overall error.

Fig.~\ref{fig:heat}(g) shows the distribution of the NTK eigenvalues along the $x$-direction in the early training stage (10,000 steps), with the data normalized. In the standard, NTK, and RAR algorithms, the eigenvalues near the boundary region and in the interior region are close in magnitude, experimentally verifying that the synchronous optimization of the standard PINN does not possess the priority of the physical information propagation path. In boundary priority, by contrast, the NTK eigenvalues decrease from the boundary toward the middle, weakening the influence of the interior points on PINN training. This indicates that the method does not merely reduce the overall error but readjusts the relative priority of the boundary neighborhood and the interior region during training: the regions near the heat-exchange boundary first obtain stronger weights, making the model preferentially form a local temperature distribution consistent with the boundary conditions; as the boundary neighborhood gradually stabilizes, the residual contribution of the interior regions is progressively released, and the boundary-induced solution structure accordingly expands into the domain, so the training process is closer to a progressive optimization path of ``constraint source $\to$ response region.''

\FloatBarrier
\subsection{Experimental Evaluation of Multi-Dimensional Priority Coupling}
\label{sec:exp_multi}

\subsubsection{Temporal--Spatial Priority Coupling: Burgers Equation}

To further evaluate the role of multi-dimensional priority coupling in temporal--spatial propagation structures, this paper selects the one-dimensional Burgers equation as a representative case. This equation simultaneously contains a nonlinear convection term and a viscous diffusion term and can describe typical phenomena such as waveform evolution, gradient steepening, and shock formation, making it suitable for verifying whether composite priority weighting, under the joint action of the temporal-advancement order and the spatial-propagation structure, can further improve PINN training on the basis of single-dimensional priority. Its governing equation can be written as:
\begin{align}
    \frac{\partial u}{\partial t} + u\frac{\partial u}{\partial x} = \nu\frac{\partial^{2} u}{\partial x^{2}}, \label{eq:burgers}
\end{align}
where $u$ denotes the fluid velocity and $\nu$ is the kinematic viscosity coefficient; the equation possesses both hyperbolic and parabolic features. In the temporal direction, the Burgers equation has two natural evolution orders: the first is the evolution order advancing from earlier states to subsequent states along the time direction, and the second is the spatial-propagation order induced by the nonlinear convection term. Based on this physical feature, this paper separately constructs a temporal priority weight and a spatial priority weight and obtains a temporal--spatial composite priority weighting through multi-dimensional multiplicative coupling. These two priorities act on different coordinate axes and, according to the directional-compatibility criterion of Chapter~3, belong to the orthogonal-compatible ($\rho = 0$) case, which can be coupled multiplicatively directly.

\begin{figure}[H]
    \centering
    \captionsetup{name=Fig.,labelfont={normalfont},labelsep=period}
    \includegraphics[width=0.76\textwidth]{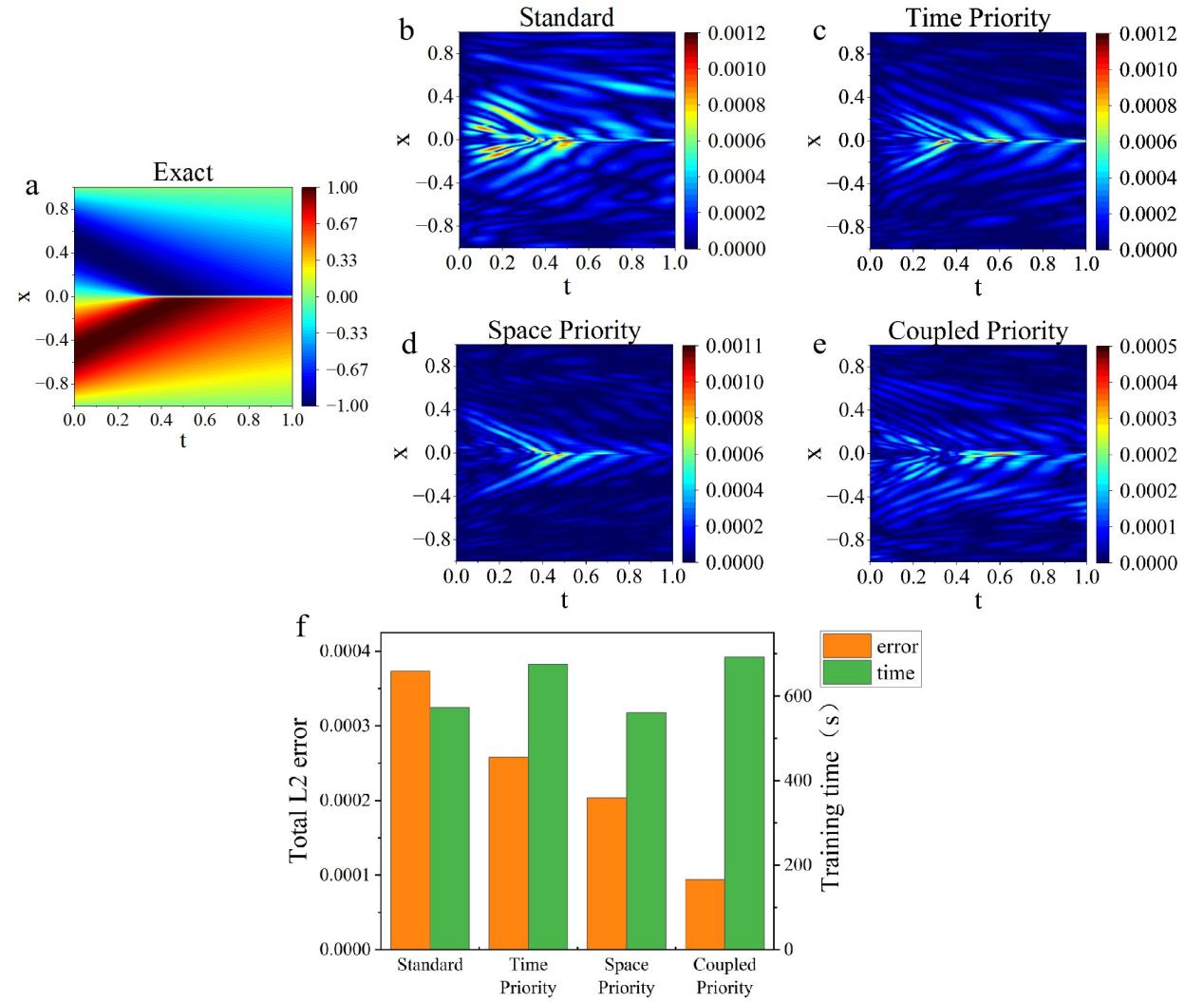}
    \caption{Reference solution of the Burgers case and the error distributions, global $L_{2}$ error, and training time under different priority-weighting strategies.}
    \label{fig:burgers}
\end{figure}

From the error contour in Fig.~\ref{fig:burgers}(b), the error of the standard PINN is concentrated mainly in regions where the solution changes drastically, especially where the gradient is large or the waveform evolves rapidly. This indicates that global synchronous training, when confronted with locally strong-variation structures in a nonlinear convection--diffusion problem, has difficulty automatically forming a training priority consistent with the solution's evolution process.

From the error contours in Fig.~\ref{fig:burgers}(c, d, e), the single-dimensional temporal priority and spatial priority can improve the convergence of the PINN from the temporal-advancement direction and the spatial-propagation direction, respectively; after the temporal--spatial composite priority is introduced, the error-concentration region is more markedly alleviated and the overall error distribution is smoother. The quantitative results in Fig.~\ref{fig:burgers}(f) further corroborate the above trend: compared with the single temporal priority and the single spatial priority, the temporal--spatial composite priority achieves a lower relative $L_{2}$ error, with the relative error reduced by 64\% relative to the temporal priority and by 54\% relative to the spatial priority. This indicates that the temporal priority and the spatial priority are complementary in the Burgers equation.

In summary, for a nonlinear PDE that simultaneously has temporal-evolution and spatial-transport features, constructing a training priority along only a single dimension is still insufficient to fully characterize the formation of the solution; incorporating both the temporal-advancement order and the spatial-propagation structure into the loss aggregation can more effectively organize the contribution of the residuals of different regions to the parameter update.

\FloatBarrier
\subsubsection{Temporal--Boundary Priority Coupling: Allen--Cahn Equation}

To further evaluate the role of multi-dimensional priority coupling in temporal--boundary propagation structures, this paper selects the Allen--Cahn equation as a representative case, whose governing equation can be written as:
\begin{align}
    \frac{\partial \phi}{\partial t} = M\varepsilon^{2}\nabla^{2}\phi - f'(\phi), \label{eq:ac}
\end{align}
where $\phi$ denotes the order parameter, $M$ denotes the mobility, $\varepsilon$ denotes the interface-width parameter, and $f'(\phi)$ denotes the nonlinear reaction term. This equation simultaneously contains a temporal-evolution process and a boundary-constraint effect: on the one hand, the phase-field structure evolves continuously in time, naturally corresponding to the ``past $\to$ future'' temporal-advancement order; on the other hand, the boundary conditions continuously constrain the admissible form of the phase field within the spatial domain and influence the evolution of the interface structure inside the domain. Therefore, this case is suitable for examining the coupling effect of the temporal priority and the boundary priority in orthogonal directions.

Based on the above space-time structure, this paper separately constructs a temporal priority weight and a boundary priority weight and obtains a temporal--boundary composite priority weight through multiplicative coupling. On the basis of the Burgers case, this case further analyzes the distribution of the NTK eigenvalues during training to further verify the effectiveness of the algorithm.

\begin{figure}[H]
    \centering
    \captionsetup{name=Fig.,labelfont={normalfont},labelsep=period}
    \includegraphics[width=0.76\textwidth]{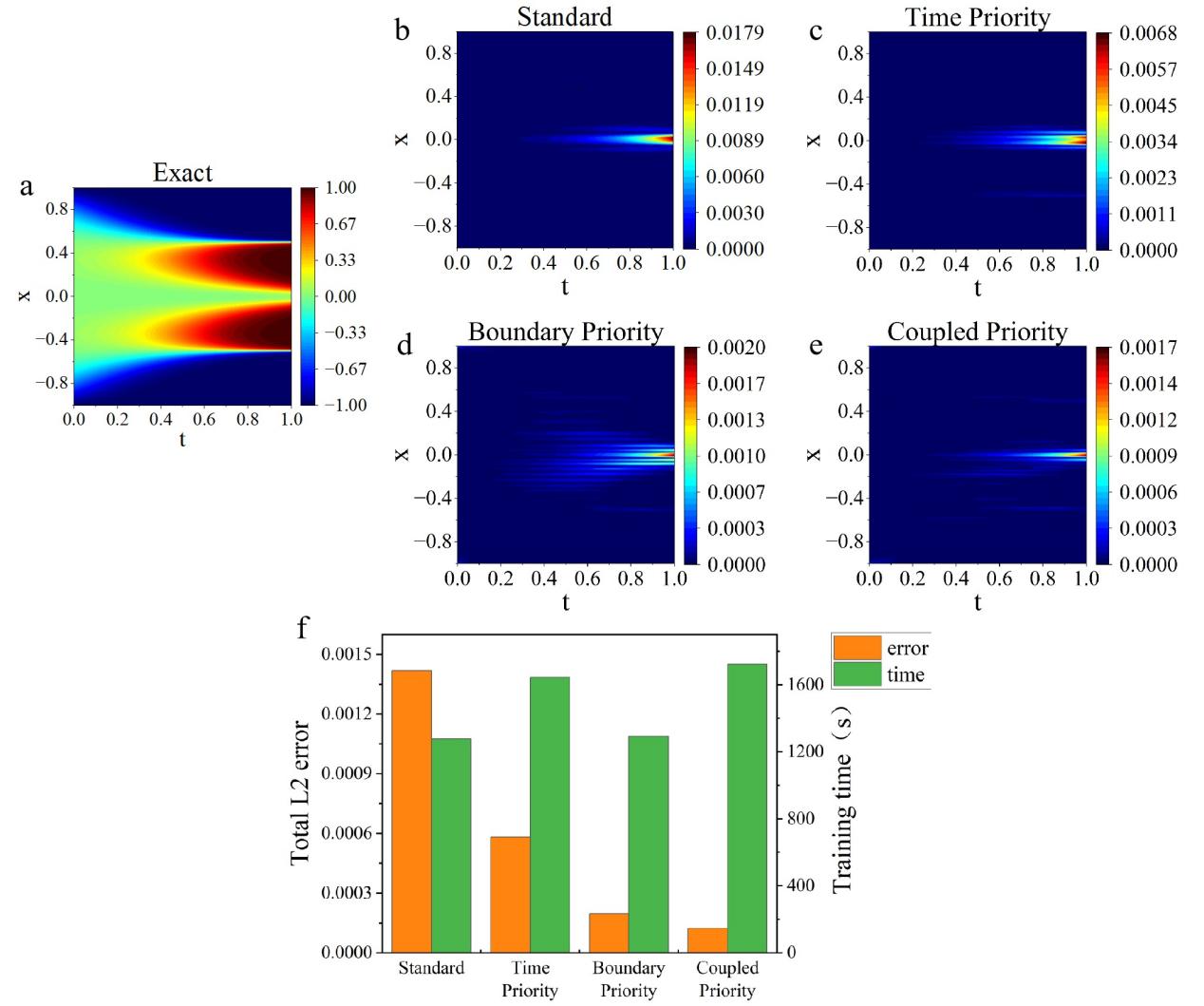}
    \caption{Reference solution of the Allen--Cahn case and the error distributions, global $L_{2}$ error, and training time under different priority-weighting strategies.}
    \label{fig:allencahn}
\end{figure}

Compared with the standard PINN, both the single temporal priority and the single boundary priority can improve the error distribution to a certain extent, further verifying the feasibility of the single-dimensional priority weights described above. Compared with the single temporal priority and the single boundary priority, the temporal--boundary composite priority achieves a lower relative $L_{2}$ error, with the relative error reduced by 79\% relative to the temporal priority and by 37\% relative to the boundary priority. When the two are further coupled, the error-concentration region is more markedly alleviated, indicating that the temporal-advancement order and the boundary-constraint priority are complementary in this problem.

\begin{figure}[H]
    \centering
    \captionsetup{name=Fig.,labelfont={normalfont},labelsep=period}
    \includegraphics[width=0.52\textwidth]{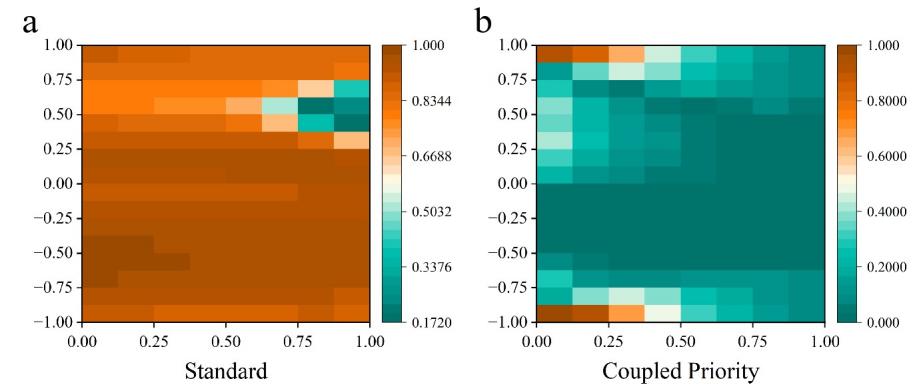}
    \caption{Heat map of the normalized NTK eigenvalue distribution in the early training stage (10,000 steps). The standard PINN shows roughly uniform eigenvalues, while the composite priority exhibits a clear decreasing structure along both the boundary and temporal dimensions.}
    \label{fig:ntk_heatmap}
\end{figure}

The results in Fig.~\ref{fig:ntk_heatmap} further support the coupling theory: in Fig.~\ref{fig:ntk_heatmap}(a), the NTK eigenvalues of the standard group are roughly uniformly distributed over the entire physical field, except for a small portion whose eigenvalues are low owing to training randomness, indicating that the synchronous optimization of the standard PINN violates the priority of the physical information propagation path in both the $x$- and $t$-directions. In the composite priority, by contrast, the NTK eigenvalues decrease not only from the boundary toward the middle but also from earlier times to later times; the algorithm simultaneously weakens the influence of later times and interior points on PINN training, prompting the PINN to learn the solution at earlier times and near the boundary first. This indicates that the multiplicative-coupling method can prompt the PINN to obey the priorities in both the boundary and temporal dimensions, thereby producing higher training accuracy.

\FloatBarrier
\subsection{Experimental Evaluation of Conflicting Multi-Dimensional Priority Coupling}
\label{sec:exp_conflict}

\subsubsection{Modeling of Complete Directional Conflict: Poiseuille Flow}

To evaluate the coupling principle when multiple propagation priorities conflict, this paper selects Poiseuille Flow as a representative case. This problem describes the stable laminar flow of an incompressible Newtonian fluid driven by a pressure gradient in a parallel-plate channel. Under the fully developed assumption, the mainstream velocity varies mainly along the wall normal, and the governing equations can be simplified to:
\begin{align}
    \frac{\partial^{2} u}{\partial y^{2}} - \frac{1}{\nu}\frac{dp}{dx} = 0,\quad \frac{\partial u}{\partial x} = 0, \label{eq:poiseuille}
\end{align}
where $u$ denotes the streamwise velocity, $\nu$ denotes the kinematic viscosity coefficient, and $dp/dx$ denotes the streamwise pressure gradient. A parabolic velocity-profile constraint is imposed at the outlet, and no-slip boundary conditions are imposed on the upper and lower walls.

Based on this physical feature, this case has two mutually opposite training priorities simultaneously in the $x$ direction. On the one hand, the mainstream direction induces a spatial priority advancing from the inlet side toward the outlet side, i.e., the ``upstream $\to$ downstream'' spatial-propagation order; on the other hand, the velocity profile at the outlet, as a strong boundary constraint source, raises the boundary priority of the outlet neighborhood in the loss function. If the boundary priority is constructed along the outlet normal, its direction is opposite to the spatial priority of the mainstream direction. Therefore, this problem constitutes a completely conflicting multi-dimensional priority situation along the same coordinate direction (the $x$ direction), corresponding to $\rho = -1$ according to the directional-compatibility criterion of Chapter~3. Accordingly, this case constructs two types of composite form: one performs multiplicative coupling in the non-conflicting direction, i.e., multiplying the selected $x$-direction priority by the $y$-direction wall-boundary priority, to evaluate the synergy among priorities of different dimensions; the other multiplicatively couples the opposite priorities in the same direction, obtaining the $x$-direction conflicting composite priority, used to examine the effect of conflicting weights on network training.

\begin{figure}[H]
    \centering
    \captionsetup{name=Fig.,labelfont={normalfont},labelsep=period}
    \includegraphics[width=0.76\textwidth]{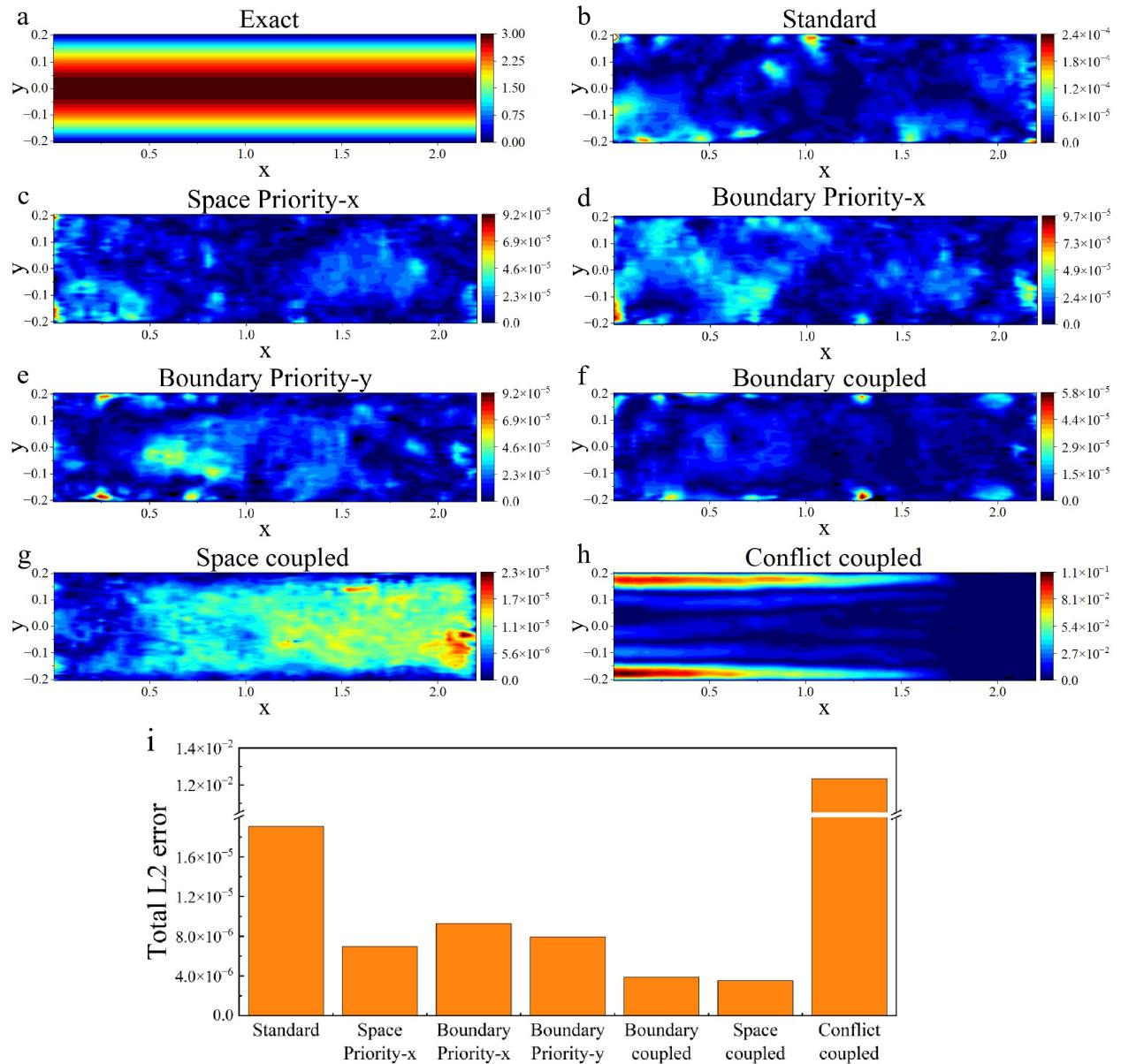}
    \caption{Error distributions and global $L_{2}$ error under different priority choices in the Poiseuille Flow case.}
    \label{fig:poiseuille}
\end{figure}

As can be seen from Fig.~\ref{fig:poiseuille}(f, g), when the $x$-direction priority is multiplicatively coupled with the non-conflicting $y$-direction wall-boundary priority, the error is further reduced. In sharp contrast, in Fig.~\ref{fig:poiseuille}(h) the $x$-direction conflicting-product priority causes the error to increase drastically, far exceeding the standard PINN and the other priority-weighting strategies. This result shows that multiplicatively coupling the ``upstream $\to$ downstream'' spatial priority and the ``outlet $\to$ interior'' boundary priority in the same direction imposes opposite training priorities on the same coordinate direction, so the residual weighting loses a clear progressive structure and transmits mutually competing optimization signals to the model, which not only fails to form a stronger constraint but instead disrupts the stability of the optimization process.

In summary, the Poiseuille Flow case shows that, in the case of complete directional conflict ($\rho = -1$), the multi-dimensional composite priority cannot be regarded as a direct multiplicative coupling of all weight terms; the effectiveness of priority weighting depends on the mutual independence of the directions, which is consistent with the $\rho$-based coupling strategy of Chapter~3.

\FloatBarrier
\subsubsection{Modeling of Partial Directional Conflict: Couette Flow}

To further examine the organization of priority weighting under locally conflicting propagation priorities, this paper selects Couette Flow as the second representative case. Unlike the complete conflict in Poiseuille Flow formed by the outlet constraint and the mainstream direction, the velocity field of Couette Flow is mainly determined jointly by the wall-shear driving and the upper and lower boundary constraints, and its priority relationship exhibits a coexistence of local consistency and local conflict in space. This paper considers a fully developed incompressible viscous shear flow between two parallel plates, with the lower wall kept stationary and the upper wall moving at a constant velocity along the $x$ direction. Considering only the mainstream velocity component, the governing equations are:
\begin{align}
    \mu\frac{d^{2} u}{d y^{2}} = 0,\quad \frac{d u}{d x} = 0, \label{eq:couette}
\end{align}
where $u$ denotes the streamwise velocity and $\mu$ is the viscosity of the fluid. A stationary condition is taken at the lower wall and a moving-wall velocity at the upper wall; since this problem is a fully developed shear flow, this paper does not impose inlet or outlet boundary conditions in the $x$ direction.

Unlike the approximately global conflict in Poiseuille Flow caused by opposite propagation priorities, the training priorities in Couette Flow exhibit more pronounced local compatibility and local competition, with two types of priority in the $y$-direction. One is the spatial priority induced by the dominant physical mechanism inside the flow field, which mainly describes the transmission order of the shear action along the channel normal, producing a $+y$-direction spatial priority advancing from the moving-wall region toward the interior together with a reverse priority produced by the reaction force; the other is the boundary priority induced by the wall boundary conditions, where the upper and lower walls, as the locations at which the boundary conditions are imposed, each induce an extension from the wall neighborhood toward the interior.

\begin{figure}[H]
    \centering
    \captionsetup{name=Fig.,labelfont={normalfont},labelsep=period}
    \includegraphics[width=0.76\textwidth]{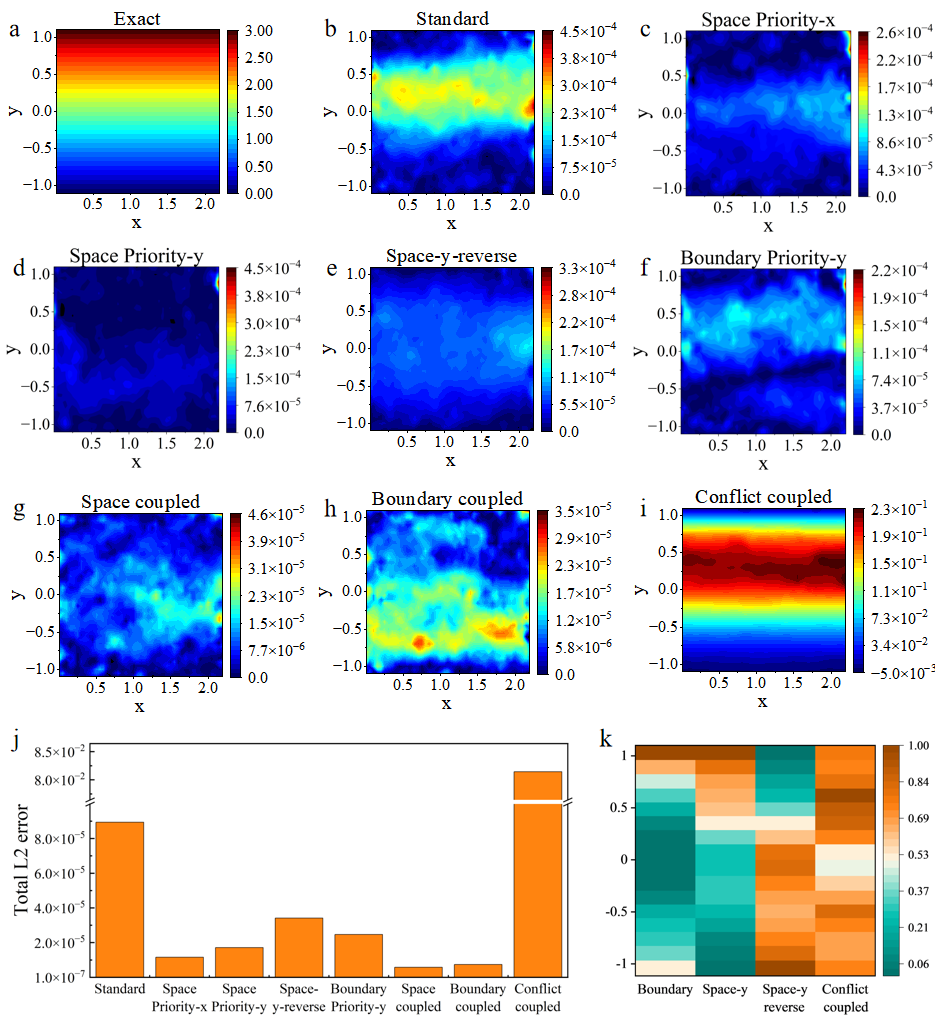}
    \caption{Reference solution of the Couette Flow case, the error distributions and global $L_{2}$ error under different priority choices and composite strategies, and the heat map of the normalized block-wise NTK eigenvalues along the $y$-axis in the early training stage (10,000 steps).}
    \label{fig:couette}
\end{figure}

From the experimental results with single-dimensional priorities introduced, each type of single priority can reduce the computational error of the standard PINN to a different extent, indicating that different single-dimensional training priorities can all provide a certain structural guidance for learning the Couette velocity profile. As can be seen from Fig.~\ref{fig:couette}(g, h), when the $x$-direction spatial priority is multiplicatively coupled with the mutually independent $y$-direction priority, the model error is further reduced. This indicates that although the main variation of Couette Flow occurs in the $y$ direction, the $x$-direction spatial priority can still serve as an auxiliary residual-organization method in the non-conflicting direction, complementing the $y$-direction priority.

From Fig.~\ref{fig:couette}(i), the $y$-conflicting-product priority imposes opposite training priorities simultaneously along the same normal direction, causing the relative $L_{2}$ error to increase significantly, far exceeding the standard PINN. In the NTK-eigenvalue heat map of Fig.~\ref{fig:couette}(k), the priorities along the boundary, $+y$, and $-y$ directions can each construct an effective training path, but the reversely coupled priority does not distinguish the magnitudes of the eigenvalues among different regions, leading to disorder in training. This result indicates that directly multiplying conflicting priorities does not form a stronger constraint, but instead makes the loss weighting lose a clear progressive structure, introduces competing optimization signals to the model, and disrupts the learning process of the velocity field.

The Couette Flow case further shows that the effectiveness of priority coupling under locally conflicting conditions depends on the mutual independence of the training priorities: composite priority weights in mutually independent directions can bring better results, whereas the direct product of opposite priorities in the same direction disrupts the stability of network training. This result is consistent with the failure phenomenon of the $x$-conflicting-product priority in Poiseuille Flow, indicating that multi-dimensional priority coupling should not be understood as a simple superposition of weight terms, but should be selectively constructed around the dominant solution-formation mechanism, i.e., the coupling principle given by the directional-compatibility criterion of Chapter~3.

The conflicting cases of the Poiseuille and Couette flows together show that the multiplicative coupling of priorities in opposite directions ($\rho = -1$) does not form a stronger constraint, but instead makes the loss weighting lose its clear progressive structure, introduces competing optimization signals to the model, and disrupts the learning process of the velocity field. This result is consistent with the directional compatibility criterion of Section~\ref{sec:multi_dim}: the applicability of multiplicative coupling depends on the mutual independence of the training priorities ($\rho = 0$), whereas the direct product of opposite priorities ($\rho = -1$) disrupts the stability of network training. Multi-dimensional priority coupling is not a simple superposition of weight terms, but should be selectively constructed around the dominant solution-formation mechanism, i.e., the coupling principle given by the directional compatibility criterion.

\section{Conclusions and Discussion}
\label{sec:conclusion}

This paper re-examines the synchronous-optimization mechanism of PINNs from the perspective of consistency with the physical information propagation path. The standard PINN incorporates the initial conditions, boundary conditions, and interior residuals in parallel into a unified loss function; for PDE problems with clear temporal-advancement, spatial-propagation, or boundary-constraint-transmission features, such global synchronous optimization is often inconsistent with the order in which the solution forms progressively ``from source to response'' along the information propagation path. Accordingly, this paper defines a unified class of training priorities: along the propagation path, premise regions should be learned before their dependent regions, and the temporal, spatial, and boundary priorities are instances of this principle on different propagation paths. Using neural tangent kernel (NTK) training dynamics, this paper further analyzes why the residual convergence order of the standard PINN is governed by the NTK spectrum, is independent of the propagation path, and is directionally biased toward the dependent regions, thereby showing that the standard PINN does not actively obey the above priority, which provides a basis for introducing explicit priority guidance.

On this basis, this paper proposes a unified multi-dimensional priority-weighting framework: by partitioning the computational domain in order along the propagation path and constructing the weights of the subsequent regions from the residuals of the preceding regions through a negative exponential, it converts the physical propagation order into a training priority at the loss-aggregation level. For cases in which multiple priorities coexist, this paper uses multiplicative coupling to characterize the synergy between mutually independent directions and proposes the directional compatibility coefficient $\rho$ as the criterion for the applicability of the coupling, introducing a selective strategy in the conflicting direction to avoid direct competition between opposite orderings.

Numerical experiments show that when the preset priority is consistent with the dominant propagation mechanism or constraint-transmission path of the problem, priority weighting can effectively improve the error distribution of the PINN and markedly reduce the overall prediction error: in the single-dimensional case, spatial priority alleviates the downstream error concentration of the steady Stokes flow past a cylinder, and boundary priority improves the error distribution in the boundary neighborhood of the heat-conduction problem; in the mutually independent multi-dimensional case, the composite priority further improves the prediction accuracy on the basis of the single dimension. The Poiseuille flow and Couette flow cases, however, reveal the applicability boundary of the coupling: direct multiplicative coupling in a conflicting direction disrupts the progressiveness of the loss weighting, introduces mutually competing optimization signals, and leads to training instability, indicating that multi-dimensional priority coupling is not a simple superposition of weight terms, but requires selecting the dominant priority in the conflicting direction and then coupling it with the non-conflicting directions.

At the same time, it should be pointed out that the effectiveness of priority weighting relies on the propagation direction or constraint-source structure being sufficiently clear; caution is still required for problems with strong global coupling, indistinct propagation directions, or a high degree of mixing of multiple mechanisms, and the exponential weights are relatively sensitive to the residual scale and the partitioning scheme, while multi-dimensional coupling also entails a certain amount of weight-construction and computational overhead. Future work may further explore automatic identification of propagation paths, adaptive weight construction, and the application of this framework to high-dimensional PDEs, multiphysics coupling, and inverse problems. In general, the framework proposed in this paper does not rely on changes to the network architecture but, by reconstructing the priority with which physical constraints enter the optimization process, improves the training stability, error distribution, and prediction accuracy of PINNs experimentally, providing a new loss-modulation approach for organizing the training of physical information in PINNs.

\appendix
\section{Details of the Experimental Configuration}
\label{sec:appendix}

All experiments were conducted on the same hardware and system; the specific specifications are given in Table~\ref{tab:hardware}. Table~\ref{tab:hyperparams} summarizes the architecture and training hyperparameters shared and differing among the six partial differential equations.

\begin{table}[htbp]
\centering
\small
\caption{Hardware and system specifications.}
\label{tab:hardware}
\begin{tabularx}{\textwidth}{@{}lX@{}}
\toprule
Component & Specification \\
\midrule
Processor & Intel Core Ultra9 185H \\
System memory & 16GB DDR5 \\
GPU & NVIDIA RTX 4060 Laptop \\
Number of GPUs & 1 \\
OS version & Ubuntu-24.04 \\
CUDA version & 12.6 \\
Performance mode & Balanced \\
\bottomrule
\end{tabularx}
\end{table}

\begin{table}[htbp]
\centering
\small
\caption{Architecture and training hyperparameters for the six PDEs.}
\label{tab:hyperparams}
\begin{tabularx}{\textwidth}{@{}lXXXXXX@{}}
\toprule
Item & Stokes & Heat & Burgers & Allen--Cahn & Poiseuille & Couette \\
\midrule
Network architecture & Modified MLP & \multicolumn{5}{c}{MLP} \\
Number of layers & 4 & 3 & 3 & 4 & 3 & 3 \\
Neurons per layer & 256 & 128 & 64 & 256 & 128 & 128 \\
Activation function & \multicolumn{6}{c}{$\tanh$} \\
Optimizer & \multicolumn{6}{c}{Adam} \\
LR decay steps & \multicolumn{6}{c}{2000} \\
Training steps & \multicolumn{6}{c}{200000} \\
Loss balancing & \multicolumn{6}{c}{Grad Norm (NTK for the NTK group)} \\
Batch size & 512 & 256 & 512 & 1024 & 256 & 256 \\
\bottomrule
\end{tabularx}
\end{table}

\end{document}